\documentclass{article}
\usepackage{spconf,amsmath,graphicx}
\usepackage{times}
\usepackage{latexsym}
\usepackage{mathtools}
\usepackage{graphicx}
\usepackage{float}
\usepackage{amsmath}
\DeclareMathOperator*{\argmax}{arg\,max}

\title{Graph-based Deep-Tree Recursive Neural Network (DTRNN) for Text Classification}
\name{Fenxiao Chen, Bin Wang and C.-C. Jay Kuo
}
\address{University of Southern California, Los Angeles, California, USA
}
\begin{document}
%
\maketitle
\begin{abstract}
A novel graph-to-tree conversion mechanism called the deep-tree
generation (DTG) algorithm is first proposed to predict text data
represented by graphs. The DTG method can generate a richer and more
accurate representation for nodes (or vertices) in graphs.  It adds
flexibility in exploring the vertex neighborhood information to better
reflect the second order proximity and homophily equivalence in a graph.
Then, a Deep-Tree Recursive Neural Network (DTRNN) method is presented
and used to classify vertices that contains text data in graphs.  To
demonstrate the effectiveness of the DTRNN method, we apply it to three
real-world graph datasets and show that the DTRNN method outperforms
several state-of-the-art benchmarking methods. 
\end{abstract}
\section{Introduction}\label{sec:intro}

Research on natural languages in graph representation has gained more
interests because many speech/text data in social networks and other
multi-media domains can be well represented by graphs.  These 
data often come in high-dimensional irregular form which makes them
more difficult to analyze than the traditional low-dimensional corpora data.
Node (or vertex) prediction is one of the most important tasks in graph
analysis. Predicting tasks for nodes in a graph deal with assigning
labels to each vertex based on vertex contents as well as link
structures.  Researchers have proposed different techniques to solve
this problem and obtained promising results using various machine
learning methods.  However, research on generating an effective
tree-structure to best capture connectivity and density of nodes in a
network is still not yet extensively conducted. 

\begin{figure*}[!ht]
\centerline{\includegraphics[width=17cm,height=6cm]{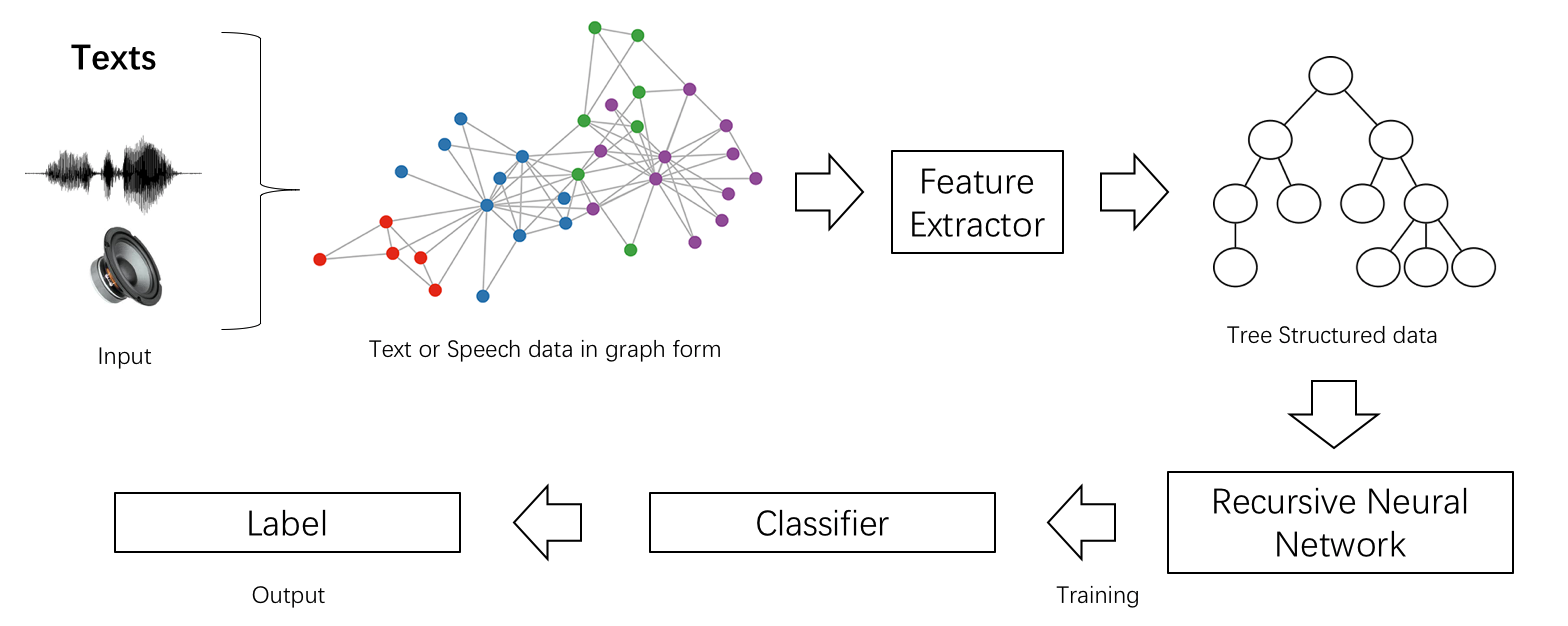}}
\caption{The workflow of the DTRNN algorithm.}\label{fig:0}
\end{figure*}

In our proposed architecture, the input text data come in form of
graphs. Graph features are first extracted and converted to tree
structure data using our deep-tree generation (DTG) algorithm. Then, the
data is trained and classified using the deep-tree recursive neural
network (DTRNN). The process generates a class prediction for each
node in the graph as the output. The workflow of the DTRNN algorithm is
shown in Figure \ref{fig:0}. 

There are two major contributions of this work. First, we propose a
graph-to-tree conversion mechanism and call it the DTG algorithm. The
DTG algorithm captures the structure of the original graph well,
especially on its second order proximity.  The second-order proximity
between vertices is not only determined by observed direct connections
but also shared neighborhood structures of vertices \cite{tang2015line}.
To put it another way, nodes with shared neighbors are likely to be
similar. Next, we present the DTRNN method that brings the merits of the
Long Short-Term Memory (LSTM) network \cite{hochreiter1997long} and the
deep tree representation together.  The proposed DTRNN method not only
conserves the link feature better but also includes the impact feature
of nodes with more outgoing and incoming edges. It extends the
tree-structured RNN and models the long-distance vertex relation on more
representative sub-graphs to offer the state-of-the-art performance as
demonstrated in our conducted experiments. An in-depth analysis on the
impact of the attention mechanism and runtime complexity of our
method is also provided. . 

The rest of this paper is organized as follows. Related previous work is
reviewed in Sec.  \ref{sec:format}. Both the DTRNN algorithm and the DTG
algorithm are described in Sec. \ref{sec:DTRNN}.  The impact of the
attention model is discussed in Sec. \ref{sec:attention}. The
experimental results and their discussion are provided in Sec.
\ref{sec:experiments}. Finally, concluding remarks are given in Sec.
\ref{sec:conclusion}. 

\section{Review of Related Work}\label{sec:format}

Structures in social networks are non-linear in nature. Network
structure understanding can benefit from modern machine learning
techniques such as embedding and recursive models. Recent studies, such
as DeepWalk \cite{perozzi2014deepwalk} and node2vec
\cite{grover2016node2vec}, aim at embedding large social networks to a
low-dimensional space. For example, the Text-Associated DeepWalk (TADW)
method \cite{Yang:15} uses matrix factorization to generate structural
and vertex feature representation.  However, these methods do not fully
exploit the label information in the representation learning. As a
result, they might not offer the optimal result. 

Another approach to network structure analysis is to leverage the
recursive neural network (RNN).  The Recursive Neural Tensor Network
(RNTN) \cite{socher2013recursive} was demonstrated to be effective in
training non-linear data structures.  The Graph-based Recurrent Neural
Network (GRNN) \cite{Xu:17} utilizes the RNTN based on local sub-graphs
generated from the original network structure.  These sub-graphs are
generated via breadth-first search with a depth of at most two. Later,
the GRNN is improved by adding an attention layer in the Attention
Graph-based Recursive Neural Network (AGRNN) \cite{xu:17att}.  Motivated
by the GRNN and AGRNN models, we propose a new solution in this work,
called the Deep-tree Recursive Neural Network (DTRNN), to improve the
node prediction performance furthermore. 

\begin{figure*}[!ht]
\includegraphics[width=\textwidth,height=5.7cm]{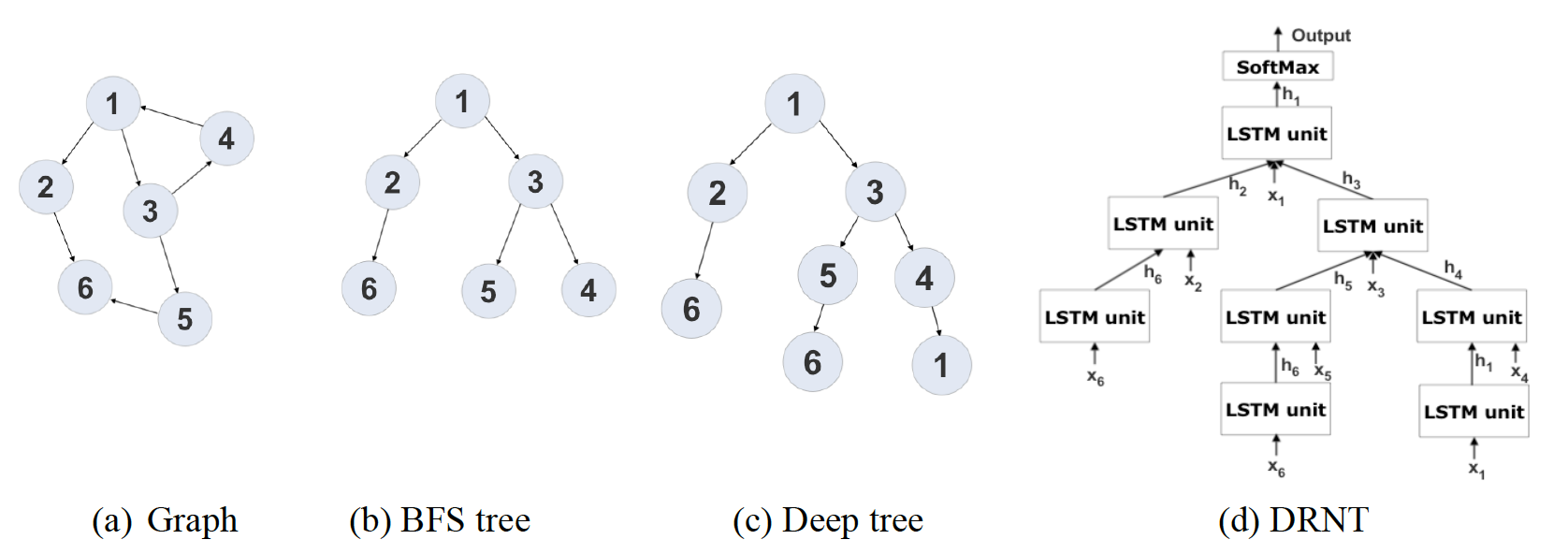}
\caption{(a) The graph to be converted into a tree, (b) the tree
converted to a sub-graph using breadth-first search, (c) the tree
converted using the deep-tree generation method (our method), and 
(d) the DTRNN constructed from the tree with LSTM units.}\label{fig:example}
\end{figure*}

\section{Proposed Methodology}\label{sec:DTRNN}

\subsection{Deep-Tree Recursive Neural Network (DTRNN) Algorithm}

A graph denoted by $G=(V,E)$ consists of a set of vertices, $V =
\{v_1,v_2,...,v_n\}$, and a set of edges, $E = \{e_{i,j}\}$, where edge
$e_{i,j}$ connects vertex $v_i$ to vertex $v_j$. Let $X_i =
\{x_1,x_2,...,x_n\}$ be the feature vector associated with vertex $v_i$,
$l_i$ be the label {or class} that $v_i$ is assigned to and $L$ the set
of all labels. The node prediction task attempts to find an appropriate
label for any vertex $v_t$ so that the probability of given vertex $v_k$
with label $l_k$ is maximized. Mathematically, we have
\begin{equation}
\hat{l_k} = \argmax_{l_k \in L} P_\theta(l_k|v_k,G).
\end{equation}
A softmax classifier is used to predict label $l_k$ of 
the target vertex $v_k$ using its hidden states $h_k$ 
\begin{equation}
P_\theta(l_k|v_k,G) =  {\rm softmax}(W_s h_k + b_s)
\end{equation}
where $\theta$ denotes model parameters.  Typically, the negative log
likelihood criterion is used as the cost function. For a network of $N$
vertices, its cross-entropy is defined as
\begin{equation}
J(\theta) = -\frac{1}{N} \sum_{k=1}^{N} \log P_{\theta}(l_k|v_k,G).
\end{equation}

To solve the graph node classification problem, we use the Child-Sum Tree-LSTM
\cite{tai2015improved} data structure to represent the node and link
information in a graph. Based on input vectors of target vertex's child
nodes, the Tree-LSTM generates a vector representation for each target
node in the dependency tree.  Like the standard LSTM, each node $v_k$
has a forget gate, denoted by $f_{kr}$, to control the memory flow
amount from $v_k$ to $v_r$; input and output gates $i_k$ and $o_k$,
where each of these gates acts as a neuron in the feed-forward neural
network, $f_{kr}$ $i_k$ and $o_k$ represent the activation of the forget
gate, input and output gates, respectively; hidden states $h_k$ for
representation of the node (output vector in the LSTM unit, and memory
cells); $c_k$ that indicates the cell state vector.  Each child takes on
input $x_k$ which is the vector representation of child nodes. 

As a result, the DTRNN method can be summarized as:
\begin{subequations}
\begin{align}
\hat{h_k} &= \max\{h_r\},\\
f_{kr} &= \sigma(W_f x_k + U_f h_c + b_f), \\
i_k &= \sigma(W_i x_k + U_i \hat{h_k}  + b_i), \\
o_k &= \sigma(W_o x_k + U_o \hat{h_k}  + b_o), \\
u_k &= \tanh(W_u x_k + U_u \hat{h_k}  + b_u), \\
c_k &= i_k\circ u_k + \sum\limits_{v_r\in C(v_k)} f_{kr} \circ c_r, \\
h_k &= o_k \circ \tanh(c_k).
\end{align}
\end{subequations}
In the equations above, $\circ$ and $\sigma$ denote the element-wise multiplication
and the sigmoid function, $W_f$, $W_i$, $W_o$ are the weights between
the forget layer and forget gate, the input layer and the input gate,
the forget gate and the output gate; $U_f$, $U_i$, $U_o$ are the weights
between the hidden recurrent layer and the forget gate, the input gate
and the output gate of the memory block; $b_f$, $b_i$, $b_o$ are the
additive biases of the forget gate, the input gate and the output gate,
respectively. 

The DTRNN is trained with back propagation through time
\cite{werbos1990backpropagation}. The model parameters are randomly
initialized. In the training process, the weight are updated after the
input has been propagated forward in the network. The error is
calculated using the negative log likelihood criterion. 

\subsection{Deep-Tree Generation (DTG) Algorithm}

In \cite{mac2017demographic}, a graph was converted to a tree using a
breadth-first search algorithm with a maximum depth of two. However, it
fails to capture long-range dependency in the graph so that the long
short-term memory in the Tree-LSTM structure cannot be fully utilized.
The main contribution of this work is to generate a deep-tree
representation of a target node in a graph. The generation starts at the
target/root node.  At each step, a new edge and its associated node are
added to the tree.  The deep-tree generation strategy is given in
Algorithm 1.  This process can be well explained using an example given
in Figure  \ref{fig:example}. 

\begin{table}
\begin{tabular}{lll}
  \hline
  \textbf{Algorithm 1} Deep-Tree Generation Algorithm\\
  \hline
  \textbf{Input:} $G$, $u_i$, maxCount\\
  \textbf{TreeGeneration} (Graph $G$, Node $u_i$, maxCount)\\
  \- \- \- \- \- \- Initialize walk to a queue $Q =$ [$u_i$]\\
  \- \- \- \- \- \- \textbf{While} $Q$ is not empty \textbf{and} \\
  \- \- \- \- \- \-totalNode $<$ maxCount \textbf{do}\\
  \- \- \- \- \- \- \- \- \- \- \- \- $v_c = Q$.pop()\\
  \- \- \- \- \- \- \- \- \- \- \- \- \textbf{if} $v$.child exists \textbf{then}\\
  \- \- \- \- \- \- \- \- \- \- \- \- \- \- \- \- \- \- \textbf{for} $w$ in G.outVertex(v) \textbf{do}\\
  \- \- \- \- \- \- \- \- \- \- \- \- \- \- \- \- \- \-\  add $w$ as the child of $v$\\
  \- \- \- \- \- \- \- \- \- \- \- \- \- \- \- \- \- \-\ $Q$.push($w$)\\
  \- \- \- \- \- \- \- \- \- \- \- \- \- \- \- \- \- \- \textbf{ end for}\\
  \- \- \- \- \- \- \- \- \- \- \- \- \textbf{end if}\\
  \- \- \- \- \- \- \textbf{end while}\\
  \- \- \- \- \- \- \textbf{return} T\\
  \hline
\end{tabular}
\end{table}

Currently, the most common way to construct a tree is to traverse the
graph using the breadth first search (BFS) method. The BFS method starts
at the tree root. It explores all immediate children nodes first before
moving to the next level of nodes until the termination criterion is
reached.  For the graph given in Figure \ref{fig:example}(a), it is
clear that node $v_5$ is connected to $v_6$ via $e_{5,6}$, and the
shortest distance from $v_4$ to $v_6$ is three hops; namely, through
$e_{4,1}$,$e_{1,2}$ and $e_{2,6}$. For the BFS tree construction process
as shown in Figure \ref{fig:example}(b), we see that such information is
lost in the translation. On the other hand, if we construct a tree by
incorporating the deepening depth first search, which is a depth limited
version of the depth first search \cite{goodrich2006algorithm}, as shown
in Algorithm 1, we are able to recover the connection from $v_5$ to
$v_6$ and get the correct shortest hop from $v_4$ to $v_6$ as shown in
Figure \ref{fig:example}(c).  Apparently, the deep-tree construction
strategy preserves the original neighborhood information better.  The
maximum number for a node to appear in a constructed tree is bounded by
its total in- and out-degrees. This is consistent with our intuition
that a node with more outgoing and incoming edges tends to have a higher
impact on its neighbors. 

\begin{figure*}[!ht]
\centerline{\includegraphics[width=21.6cm,height=6cm]{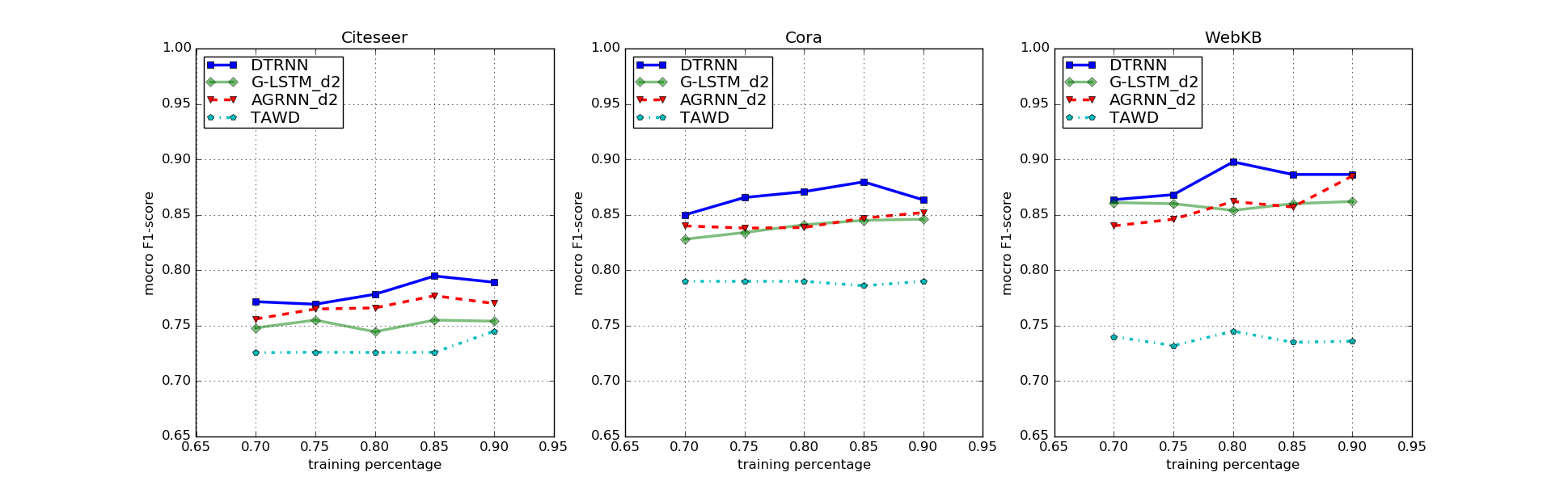}}
\caption{Comparison of four methods on three data sets (from left to
right): Citeseer, Cora and WebKB, where the x-axis is the percentage of
training data and the y-axis is the average Macro-F1 score.}\label{fig:performance}
\end{figure*}

\begin{figure*}[!ht]
\centerline{\includegraphics[width=21.6cm,height=6cm]{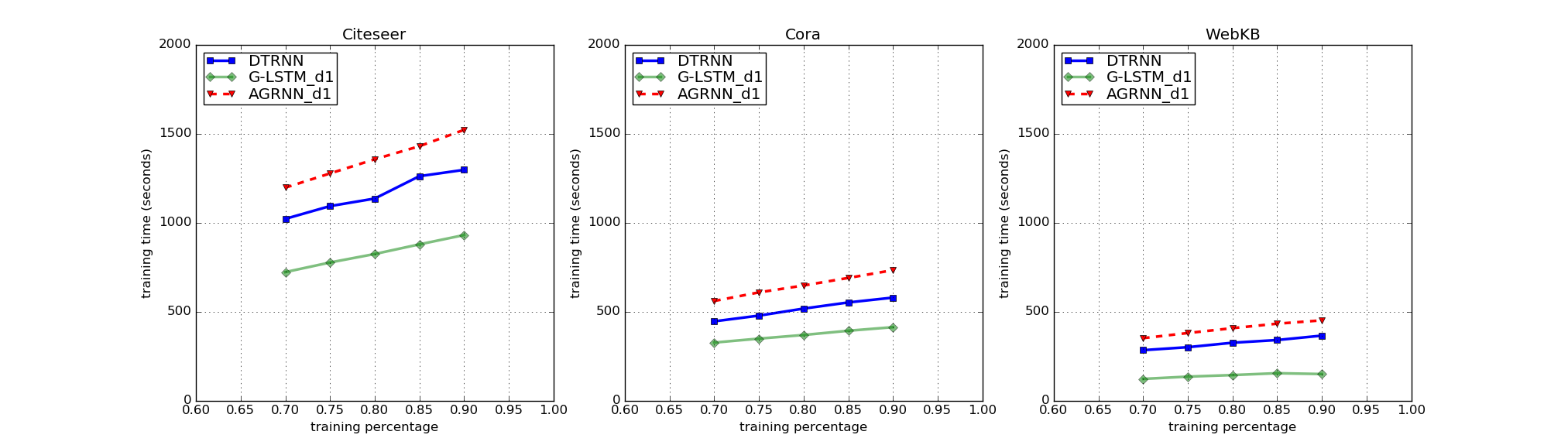}}
\caption{Comparison of runtime for three data sets (from left to
right): Citeseer, Cora and WebKB, where the x-axis is the percentage of
training data and the y-axis is the runtime in seconds.}\label{fig:4}
\end{figure*}

\begin{figure*}[!ht]
\centerline{\includegraphics[width=21.6cm,height=6cm]{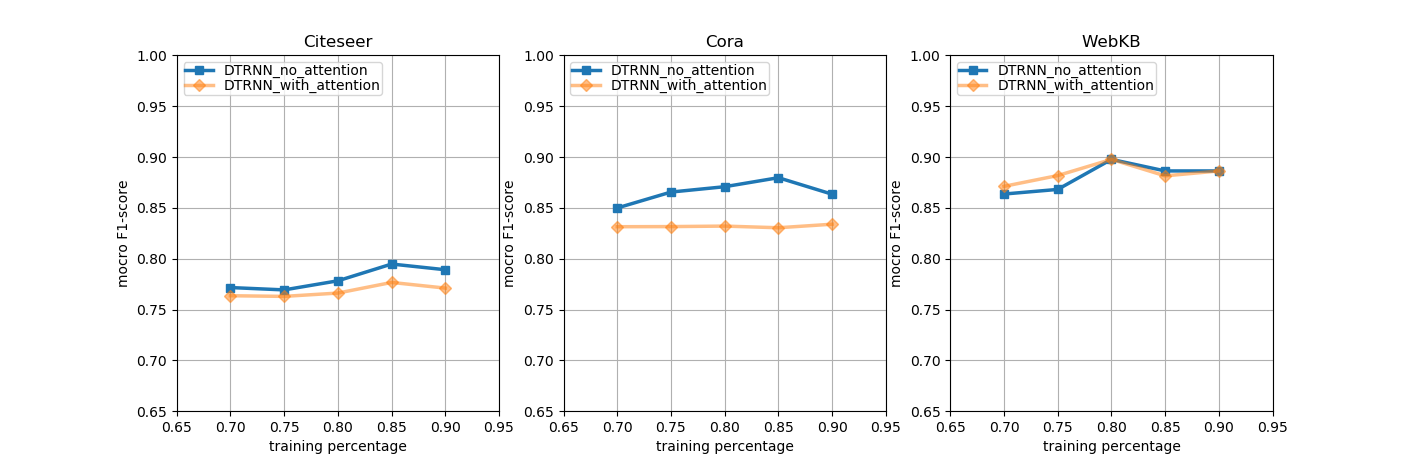}}
\caption{Performance comparison of DTRNN with and without the soft attention layer 
(from left to right: Citeseer, Cora and WebKB), where the x-axis is the percentage of
training data and the y-axis is the average Macro-F1 score.}\label{fig:performance}
\end{figure*}

\section{Impact of Attention Model}\label{sec:attention}

An attentive recursive neural network can be adapted from a regular
recursive neural network by adding an attention layer so that the new
model focuses on the more relevant input. The attentive neural network
has demonstrated improved performance in machine translation, image
captioning, question answering and many other different machine learning
fields. The added attention layer might increase the classification
accuracy because the graph data most of the time contain noise. The less
irrelevant neighbors should has less impact on the target vertex than
the neighbors that are more closely related to the target vertex.
Attention models demonstrated improved accuracy in several applications. 

In this work, we examine how the added attention layers could affect the
results of our model. In the experiment, we added an attention layer to
see whether the attention mechanism could help improve the proposed
DTRNN method.  The attention model is taken from \cite{xu:17att} that
aims to differentiate the contribution from a child vertex to a target
vertex using a soft attention layer. It determines the attention weight,
$\alpha_r$, using a parameter matrix denoted by $W_\alpha$.  Matrix
$W_\alpha$ is used to measure the relatedness of $x$ and $h_r$. It is
learned by the gradient descent method in the training process. The
softmax function is used to set the sum of attention weights to equal 1.
The aggregated hidden state of the target vertex is represented as the
summation of all the soft attention weight times the hidden states of
child vertices as
\begin{equation}\label{eq:att-1}
\alpha_r = {\rm Softmax} (x^T W_\alpha h_r),
\end{equation}
where
\begin{equation}\label{eq:att-2}
\hat{h_r} =\alpha_r h_r.
\end{equation}
Based on Eqs. 4(a), (5) and (6), we can obtain
\begin{equation}\label{eq:att-3}
\hat{h_k} = \max{\{ {\rm Softmax} (x^T W_\alpha h_r) h_r \} }.
\end{equation}
Athough the attention model can improve the overall accuracy of a
simple-tree model generated by a graph, its addition does not help 
but hurts the performance of the proposed deep-tree model. This
could be attributed to several reasons.

It is obvious to see that $\alpha_r$ is bounded between 0 and 1 because
of the softmax function. If one target root has more child nodes,
$\alpha_r$ will be smaller and getting closer to zero. By comparing
Figures \ref{fig:example}(b) and (c), we see that nodes that are further
apart will have vanishing impacts on each other under this attention
model since our trees tend to have longer paths.  The performance
comparision of DTRNN with and without attention added is given in Figure
\ref{fig:performance}. For Cora, we see that DTRNN without the attention
layer outperforms the one with attention layer by 1.8-3.7\%. For
Citeseer, DTRNN without the attention layer outperforms by 0.8-1.9\%.
For WebKB, the performance of the two are about the same. 

Furthermore, this attention model pays close attention to the immediate
neighbor of a target yet ignores the second-order proximity, which can
be interpreted as nodes with shared neighbors being likely to be similar
\cite{tang2015line}. Prediction tasks on nodes in networks should take
care of two types of similarities: (1) homophily and (2) structural
equivalence \cite{hoff2002latent}. The homophily hypothesis
\cite{yang2014overlapping} states that nodes that are highly
interconnected and belong to similar network clusters or communities
should be similar to each other.  The vanishing impact of scalded $h_r$
tends to reduce these features in our graph.  In the next section, we
will show by experiments that the DTRNN method without the attention
model outperforms a tree generated by the traditional BFS method with an
attention LSTM unit and also DTRNN method with attention model . 

\section{Experiments}\label{sec:experiments}

\subsection{Datasets}
To evaluate the performance of the proposed DTRNN method, we used the
following two citation and one website datasets in the experiment.
\vspace{-0.75em}
\begin{itemize}
\setlength\itemsep{-0.4em}
\item \textbf{Citeseer}: The Citeseer dataset is a citation indexing
system that classifies academic literature into 6 categories
\cite{Giles:98}. This dataset consists of 3,312 scientific publications
and 4,723 citations. 
\item \textbf{Cora}: The Cora dataset consists of 2,708 scientific
publications classified into seven classes \cite{McCallum:00}. This
network has 5,429 links, where each link is represented by a 0/1-valued
word vector indicating the absence/presence of the corresponding word
from a dictionary consists of 1,433 unique words. 
\item \textbf{WebKB}: The WebKB dataset consists of seven classes of web
pages collected from computer science departments: student, faculty,
course, project, department, staff and others \cite{craven1998learning}. It
consists of 877 web pages and 1,608 hyper-links between web pages. 
\end{itemize}
\subsection{Experimental Settings}
\vspace{-0.65em}
These three datasets are split into training and testing sets
with proportions varying from 70\% to 90\%. We run 10 epochs on the
training data and recorded the highest and the average Micro-F1 scores 
for items in the testing set. 

\subsection{Baselines}
We implemented a DTRNN consisting of 200 hidden states, and compare its
performance with that of three benchmarking methods, which are described
below. 
\vspace{-0.75em}
\begin{itemize}
\setlength\itemsep{-0.4em}
\item Text-associated Deep Walk (TADW). It incorporates text features of
vertices under the matrix factorization framework \cite{Yang:15} for
vertex classification. 
\item Graph-based LSTM (G-LSTM). It first builds a simple tree using the
BFS only traversal and, then, applies an LSTM to the tree for vertex
classification \cite{Xu:17}. 
\item Attentive Graph-based Recursive Neural Network (AGRNN).  It is
improved upon the GRNN with soft attention weight added in the each
attention unit as depicted in Eqs. (\ref{eq:att-1}) and (\ref{eq:att-2})
\cite{xu:17att}. 
\end{itemize}

\subsection{Results and Analysis}

The Macro-F1 scores of all four methods for the above-mentioned three
datasets are compared in Figure \ref{fig:performance}. We see that the
proposed DTRNN method consistently outperforms all benchmarking methods.
When comparing the DTRNN and the AGRNN, which has the best performance
among the three benchmarks, the DTRNN has a gain up to 4.14\%.  The
improvement is the greatest on the WebKB dataset. In the Cora and the
Citeseer datasets, neighboring vertices tend to share the same label.
In other words, labels are closely correlated among short range
neighbors.  In the WebKB datasets, this short range correlation is not
as obvious, and some labels are strongly related to more than two labels
\cite{Xu:17}. Since our tree-tree generation strategy captures the long
distance relation among nodes, we see the largest improvement in this
dataset. 

\subsection{Complexity Analysis}\label{sec:complexity}

The graph-to-tree conversion is relatively fast. For both the
breadth-first and our method, the time complexity to generate the tree
is $b^d$, where $b$ is the max branching factor of the tree, and $d$ is
the depth. The DTRNN algorithm builds a longer tree with more depth.
Thus, the tree construction and training will take longer yet  overall it still
grows linearly with the number of input node asymptotically.  

The bottleneck of the experiments was the training process. During each
training time step, the time complexity for updating a weight is $O(1)$.
Then, the overall LSTM algorithm has an update complexity of $O(W)$ per
time step, where $W$ is the number of weights \cite{hochreiter1997long}
that need to be updated.  In addition, LSTM is local in space and time,
meaning that it does not depend on the network size to update complexity
per time step and weight, and the storage requirement does not depend on
input sequence length \cite{schmidhuber1989local}. For the whole
training process, the run time complexity is $O(W i e)$, where $i$ is
the input length and $e$ is the number of epochs. 

In our experiments, the input length is fixed per time step because the
hidden states of the child vertices are represented by max pooling of
all children's inputs. The number of epochs is fixed at 10. The actual
running time for each data set is recorded for the DTRNN method and the
G-LSTM method. The results are shown in Figure 3.  If attention layers
are added as described in the earlier section, they come at a higher
cost. The attention weights need to be calculated for each combination
of child and target vertex. If we have $c$ children on average for $n$
target vertices, there will be $c \times n$ attention values.  The
actual machine runtime of three datasets are shown in Figure
\ref{fig:4}.  The CPU runtime shows that the DTRRN is faster than the
AGRNN-d1 (with an attention model of depth equal to 1) by 20.59\% for
WebKB, 14.78\% for Citeseer, and 21.06\% for Cora while having the
highest classification accuracy among all three methods. 

\section{Conclusion and Future Work}\label{sec:conclusion}

A novel strategy to convert a social citation graph to a deep tree and
to conduct the vertex classification problem was proposed in this work.
It was demonstrated that the proposed deep-tree generation (DTG)
algorithm can capture the neighborhood information of a node better than
the traditional breath first search tree generation method. Experimental
results on three citation datasets with different training ratios proved
the effectiveness of the proposed DTRNN method. That is, our DTRNN
method offers the state-of-the-art classification accuracy for graph
structured text. 

We also trained graph data in the DTRNN by adding more complex attention
models, yet attention models does not generate better accuracy because
our DTRNN algorithm alone already captures more features of each node.
The complexity of the proposed method was analyzed. We considered both
asymptotic run time and real time CPU runtime and showed that our
algorithm is not only the most accurate but also very efficient. 

In the near future, we would like to apply the proposed methodology to
graphs of a larger scale and higher diversity such as social network
data.  Furthermore, we will find a new and better way to explore the
attention model although it does not help much in our current
implementation. 

\bibliographystyle{IEEE}
\bibliography{strings,refs}

\end{document}